\documentclass[conference]{IEEEtran}
\IEEEoverridecommandlockouts
\usepackage{cite}
\usepackage{amsmath,amssymb,amsfonts,nccmath, amsthm}
\usepackage{graphicx}
\usepackage{textcomp}
\usepackage{xcolor}

\usepackage{bm}
\usepackage[noend]{algpseudocode}
\usepackage{algorithmicx}
\usepackage{algorithm}
\usepackage{multirow}
\usepackage{booktabs}
\usepackage{pifont}
\usepackage{verbatim}
\usepackage{balance}
\usepackage{footmisc}

\usepackage[caption=false,font=footnotesize,labelfont=sf,textfont=sf,subrefformat=parens,labelformat=parens]{subfig}

\def\BibTeX{{\rm B\kern-.05em{\sc i\kern-.025em b}\kern-.08em
    T\kern-.1667em\lower.7ex\hbox{E}\kern-.125emX}}

\begin{document}

\title{FedGroup: Efficient Federated Learning via Decomposed Similarity-Based Clustering
}

\author{\IEEEauthorblockN{Moming Duan\IEEEauthorrefmark{1}, 
		Duo Liu\IEEEauthorrefmark{1},
		Xinyuan Ji\IEEEauthorrefmark{2},
		Renping Liu\IEEEauthorrefmark{1},
		Liang Liang\IEEEauthorrefmark{1},
		Xianzhang Chen\IEEEauthorrefmark{1},
		Yujuan Tan\IEEEauthorrefmark{1}}
	\IEEEauthorblockA{\IEEEauthorrefmark{1}College of Computer Science, Chongqing University, Chongqing, China\\
	}
	\IEEEauthorblockA{\IEEEauthorrefmark{2}School of Computer Science and Technology, Xi'an Jiaotong University, Xi'an, China\\
	}
}

\maketitle

\begin{abstract}
Federated Learning (FL) enables the multiple participating devices to collaboratively contribute to a global neural network model while keeping the training data locally. 
Unlike the centralized training setting, the non-IID and imbalanced (statistical heterogeneity) training data of FL is distributed in the federated network, which will increase the divergences between the local models and the global model, further degrading performance. 
In this paper, we propose a novel clustered federated learning (CFL) framework FedGroup, in which we 
1) group the training of clients based on the similarities between the clients' optimization directions for high training performance; 
2) construct a new data-driven distance measure to improve the efficiency of the client clustering procedure.
3) implement a newcomer device cold start mechanism based on the auxiliary global model for framework scalability and practicality.

FedGroup can achieve improvements by dividing joint optimization into groups of sub-optimization and can be combined with FL optimizer \textit{FedProx}.
The convergence and complexity are analyzed to demonstrate the efficiency of our proposed framework.
We also evaluate FedGroup and FedGrouProx (combined with \textit{FedProx}) on several open datasets and made comparisons with related CFL frameworks. The results show that FedGroup can significantly improve absolute test accuracy by $+14.1\%$ on FEMNIST compared to \textit{FedAvg}, $+3.4\%$ on Sentiment140 compared to \textit{FedProx}, $+6.9\%$ on MNIST compared to \textit{FeSEM}.

\end{abstract}


\section{Introduction}

Federated Learning (FL)~\cite{konevcny2016federated,mcmahan2017communication,bonawitz2019towards,yang2019federated,li2020federated} is a promising distributed neural network training approach, which enables multiple end-users to collaboratively train a shared neural network model while keeping the training data decentralized. 
In practice, a FL server first distributes the global model to a random subset of participating clients (e.g. mobile and IoT devices). 
Then each client optimizes its local model by gradient descent based on its local data in parallel. 
Finally, the FL server averages all local models' updates or parameters and aggregates them to construct a new global model.
Unlike the traditional cloud-centric learning paradigm and the distributed machine learning frameworks based on Parameter Server~\cite{li2014scaling}, there is no need to transfer private data over the communication network during the FL training.
With the advantage of privacy-preserving, Federated Learning is currently the most attractive distributed machine learning framework.
Nevertheless, due to the FL server does not have the authority to access the user data or collect statistical information, some data preparation operations such as balancing and outlier detection be restricted. 
Therefore, the high statistical heterogeneity is a challenging problem in federated learning~\cite{li2020federated}.

To tackle heterogeneity in federated learning, several efforts have been made.
McMahan \textit{et al.} propose the vanilla FL framework \textit{Federated Averaging (FedAvg)}~\cite{mcmahan2017communication} and experimentally demonstrate that \textit{FedAvg} is communication-efficient and can converge under statistical heterogeneity setting (non-IID). 
However, Zhao \textit{et al.}~\cite{zhao2018federated} show that the accuracy reduces $\sim$55\% for CNN trained on highly skewed CIFAR-10~\cite{krizhevsky2009learning}. 
The experiments based on VGG11~\cite{Simonyan15} by Sattler \textit{et al.}~\cite{sattler2019robust} show that non-IID data not only leads to accuracy degradation, but also reduces the convergence speed. 
Li \textit{et al.}\cite{Li2020On} theoretically analyze the convergence of \textit{FedAvg} and indicates that the heterogeneity of data slows down the convergence for the strongly convex and smooth problem. 
In addition, Duan \textit{et al.}~\cite{duan2019astraea} demonstrate that global imbalanced data also has adverse effects on federated training. 
Unfortunately, the retrieval of model accuracy decreases as the local model diverges in~\cite{zhao2018federated,sattler2019robust,duan2019astraea}. 
Recently, Sattler \textit{et al.}~\cite{sattler2020clustered, sattler2020byzantine} propose a novel federated multi-task learning framework Cluster Federated Learning (CFL), which exploits geometric properties of FL loss surface to cluster the learning processes of clients based on their optimization direction, provides a new way of thinking about the statistical heterogeneity challenge. Many researchers follow up on CFL-based framework \cite{xie2020multi, ghosh2020efficient, briggs2020federated, zhang2021csafl} and confirm CFL is more accurate than traditional FL with the consensus global model. However, above CFL-based frameworks are inefficient in the large-scale federated training systems or ignore the presence of newcomer devices. 

In this paper, we propose an efficient and accurate clustered federated learning framework FedGroup, which clusters clients into multiple groups based on a new decomposed data-driven measure between their parameter updates. 
In each communication round, each active client only contributes its local optimization result to the corresponding group model.
The framework still maintains an auxiliary server to address the cold start issues of new devices.
To improve the performance of high-dimension low-sample size (HDLSS) parameter updates clustering, we use a novel data-driven measure of cosine dissimilarity called Euclidean distance of Decomposed Cosine similarity (EDC), which can also avoid the concentration phenomenon of $\ell_p$ distances in high dimensional data clustering~\cite{sarkar2019perfect}.
Furthermore, by combining FedGroup with the federated optimizer \textit{FedProx}~\cite{li2018federated},  FedGroup can be revised as FedGrouProx, which be explored in our experiments.

With the above methods, FedGroup can significantly improve test accuracy by $+6.9\%$ on MNIST~\cite{lecun1998gradient}, $+26.9\%$ on FEMNIST~\cite{cohen2017emnist}, $+5.3\%$ on Sentiment140~\cite{go2009twitter} compared to FedSEM. We show that FedGroup has superior performance than FedProx and FeSEM~\cite{xie2020multi}. Although FedGroup achieves performance improvements similar to IFCA~\cite{ghosh2020efficient}, the latter has more communication and time overhead. The ablation studies of FedGroup are provided to demonstrate the usefulness of our clustering and cold start strategies.

The main contributions of this paper are summarized as follows.
\begin{itemize}
	\item We propose two novel clustered federated learning frameworks, FedGroup and FedGrouProx, and show their superiority on four open datasets (with statistical heterogeneity) compared to several FL and CFL frameworks.
	\item We propose a decomposed data-driven measure named EDC to improve the effectiveness of client clustering in FedGroup, which prevents the adverse effects of concentration phenomenon caused by directly clustering HDLSS model parameters or direction vectors. 
	\item Our framework presents an efficient cold start strategy for the groups and the newcomers, which provides a new approach to improve the scalability and practicality of the previous works. We demonstrate its efficiency experimentally. In addition, we open source the code of FedGroup to contribute to the community.
\end{itemize}

The rest of this paper is organized as follows.
Section~\ref{sec:background} provides the background of FL and an overview of related works.
Section~\ref{sec:design} shows the motivation and the design of FedGroup. 
The evaluation results are presented and analyzed in Section~\ref{sec:evaluation}. 
Section~\ref{sec:conclusion} concludes the paper.


\section{background}\label{sec:background}
\subsection{Federated Learning}
In this section, we introduce the most widely adopted FL algorithm FedAvg and briefly explain the optimization method FedProx.
McMahan \textit{et al.} first introduce federated learning~\cite{mcmahan2017communication} and the vanilla FL optimization method FedAvg, which is designed to provide privacy-preserving support for distributed machine learning model training.
The distributed objective of FL is:

\begin{equation}
	\min\limits_{{\bm{w}}} \Big\{f(\bm{w}) \triangleq \sum_{k=1}^{N}p_k F_k(\bm{w})\Big\},
\end{equation}

Where $N$ is the number of clients, $p_k$ is the weight of the $k$-th device, $p_k \geqslant 0$, $\sum_{k}^{} p_k = 1$.
In statistical heterogeneity setting, the local objectives $F_k(\bm{w})$ measure the local expirical risk over possibly differing data distributions $p_{data}^{(k)}$. 

For a machine learning problem, we can set $F_k(\bm{w})$ to the user-specified loss function $L \big( \cdot;\cdot \big)$ of the predictions on examples $(\bm{x},\bm{y})$ made with model parameters $w$. Hence, the local objective is defined by

\begin{equation}
	F_k(\bm{w}) \triangleq \mathbb{E}_{(\bm{x},\bm{y})\sim{p}_{data}^{(k)}}L\big( \bm{x},\bm{y}; \bm{w} \big).
\end{equation}

The global objective $f(\bm{w})$ can be regarded as a joint objective function for multiple clients, and FL tries to minimize it by optimizing all local optimization objectives.

In practice, a typical implementation of FL system~\cite{bonawitz2019towards} includes a cloud server that maintains a global model and multiple participating devices, which communicate with the server through a network.
At each communication round $t$, the server selects a random subset $K$ of the active devices (i.e. clients) to participate in this round of training.
The server then broadcasts the latest global model $w^t$ to these clients for further local optimization.
Then each participating client optimizes the local objective function based on the device's data by its local solvers (e.g. SGD) with several local epochs $E$ in parallel.
Then, the locally-computed parameter updates $\Delta \bm{w}_i^t$ from these clients that completed the training within the time budget are collected and aggregated.
Finally, the server will update the global to $w^{t+1}$ and finish the current round.
In general, the FL training task requires hundreds of rounds to reach target accuracy.

The details of FedAvg are shown in Algorithm~\ref{alg:fedavg}. 
Here $n_i$ denotes the training data size of client $i$, the total training data size $n=\sum n_i$. The $\mathcal{B}_i$ is the batches of training data of client $i$, $\eta$ is the learning rate of the local solver. 
There have two key hyperparameters in $FedAvg$, the first is the number of participating clients $K$ in each round, or the participation rate $K/N$. 
For the IID setting, a higher participation rate can improve the convergence rate, but for the  non-IID setting, a small participation rate is recommended to alleviate the straggler's effect~\cite{Li2020On}. 
The second is the local epoch $E$, an appropriately large $E$ can increase the convergence speed of the global optimization and reduce the communication requirement. 
However, an excessively large $E$ will increase the discrepancy between the local optimization solutions and the global model, which will lead to the federated training procedure be volatile and yields suboptimal results.

\begin{algorithm}[t]
	\caption{Federated Averaging (FedAvg)}
	\label{alg:fedavg}
	\begin{algorithmic}[1]
		\Procedure {FL Server Training}{}
		\State Initialize global model $\bm{w}_0$, then $\bm{w}_1 \leftarrow \bm{w}_0$.
		\For{each communication round $t=1,2,...,T$}
			\State $S_t \leftarrow$ Server selects a random subset of $K$ clients.
			\State Server broadcasts $\bm{w}_t$ to all selected clients.
			\For{each activate client $i \in S_t$ parallelly}
				\State $\Delta\bm{w}_{t+1}^i\leftarrow$~\textbf{ClientUpdate($i$,~$\bm{w}_t$)}.
			\EndFor
			\State $\bm{w}_{t+1}\leftarrow\bm{w}_t+\sum_{i \in S_t} \frac{n_i}{n} \Delta\bm{w}_{t+1}^i$
		\EndFor
		\EndProcedure

		\Function{ClientUpdate}{$i$,~$\bm{w}$}
		\State $\hat{\bm{w}}\leftarrow\bm{w}$.
		\For{each local epoch $e=1,2,...,E$}
		\State $\bm{w}\leftarrow\bm{w}-\eta\nabla L(b;\bm{w})$ for local batch $b \in \mathcal{B}_i$.
		\EndFor
		\State \Return$\Delta\bm{w}\leftarrow\bm{w}-\hat{\bm{w}}$
		\EndFunction
	\end{algorithmic}
\end{algorithm}

As mentioned above, the key to increasing the convergence speed and robustness of the FL system is to bound the discrepancies between the local optimization objectives and the global objective. To alleviate the divergences, Li \textit{et al.}~\cite{li2018federated} introduce a penalty (aka proximal term) to the local objective function: 
\begin{equation}\label{equ:prox}
	F_k(\bm{w})+\frac{\mu}{2}\|\bm{w}-\bm{w}^t\|^2. 
\end{equation}
Where the hyperparameter $\mu$ controls how far the updated model can be from the starting model. Obviously that FedAvg is a special case of FedProx with $\mu=0$.

\subsection{Clustered Federated Learning}\label{sec:background-CLF}
One of the main challenges in the design of the large-scale FL system is statistical heterogeneity (e.g. non-IID, size imbalanced, class imbalanced)~\cite{li2020federated, bonawitz2019towards, sattler2019robust, briggs2020federated, duan2020self, li2021fedsae}. 
The conventional way is to train a consensus global model upon incongruent data, which yields unsatisfactory performance in the high heterogeneity setting~\cite{zhao2018federated}. 
Instead of optimizing a consensus global model, CFL divides the optimization goal into several sub-objectives and follows a \textit{Pluralistic Group} architecture~\cite{lee2020tornadoaggregate}. CFL maintains multiple groups (or clusters) models, which are more specialized than the consensus model and achieve high accuracy. 

Sattler \textit{et al.} propose the first CFL-based framework~\cite{sattler2020clustered}, which recursively separate the two groups of clients with incongruent descent directions. The authors further propose~\cite{sattler2020byzantine} to improve the robustness of CFL-based framework in byzantine setting.
However, the recursive bi-partitioning algorithm is computational inefficient and requires multiple communication rounds to completely separate all incongruent clients.
Furthermore, since the participating client in each round is random, this may cause the recursive bi-partitioning algorithm to fail.
To improve the efficiency of CFL, Ghosh \textit{et al.} propose IFCA~\cite{ghosh2020efficient}, which randomly generates cluster centers and divides clients into clusters that will minimize their loss values.
Although the model accuracy improvements of IFCA are significant, IFCA needs to broadcast all group models in each round and is sensitive to the initialization with probability success.
FeSEM~\cite{xie2020multi} uses a $\ell_2$ distance-based stochastic expectation maximization (EM) algorithm instead of distance-based neighborhood methods. However, $\ell_2$ distance often suffer in HDLSS situation which is known as distance concentration phenomenon in high dimension~\cite{sarkar2019perfect}.
The distance concentration of $\ell_p$ will cause the violation of neighborhood structure~\cite{radovanovic2010existence}, which has adverse effects on the performance of pairwise distance-based clustering algorithms such as K-Means, k-Medoids and hierarchical clustering~\cite{sarkar2019perfect}.
Similarity to the cosine similarity-based clustering method~\cite{sattler2020clustered}, Briggs \textit{et al.} propose an agglomerative hierarchical clustering method named FL+HC~\cite{briggs2020federated}.
It relies on iterative calculating the pairwise distance between all clusters, which is computationally complex. 
Note that, all the above CFL frameworks assume that all clients will participate in the clustering process and no newcomer devices, which is unpractical in a large-scale FL system.

\section{FedGroup and FedGrouProx}\label{sec:design}

\subsection{Motivation}
Before introducing FedGroup and FedGrouProx, we show a toy example to illustrate the motivation of our work. 
To study the impacts of statistical heterogeneity, model accuracy, and discrepancy, we implement a convex multinomial logistic regression task train with FedAvg on MNIST~\cite{lecun1998gradient} following the instructions of~\cite{li2018federated}. 
We manipulate the statistical heterogeneity by forcing each client to have only a limited number of classes of data. 
All distributed client data is randomly sub-sampled from the original MNIST dataset without replacement, and the training set size follows a power law. 

\begin{figure*}[t]
	\centering
	\includegraphics[scale=1.0]{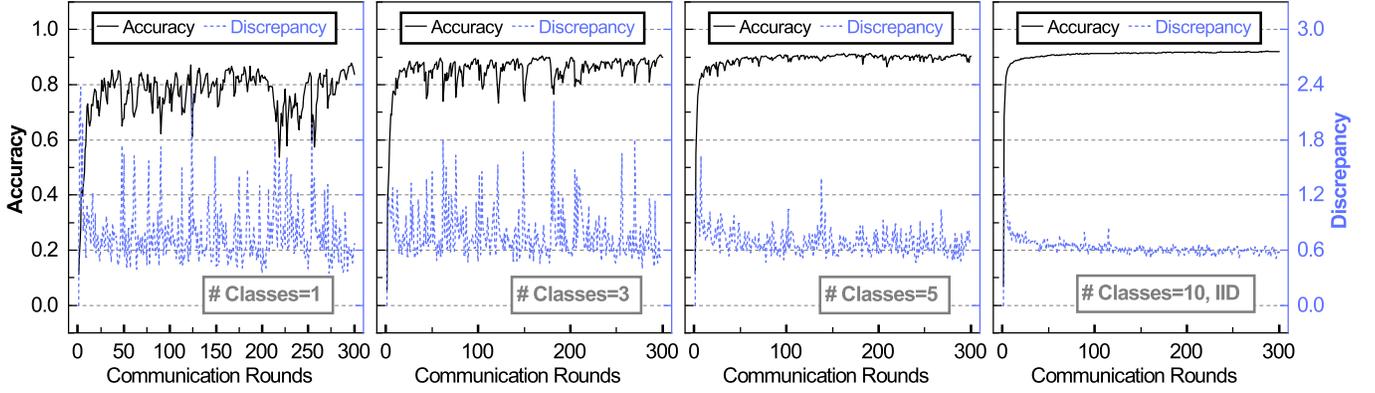}
	\caption{A FedAvg training procedure on three non-IID MNIST datasets and one IID MNIST dataset to illustrate the effects of statistical heterogeneity on model accuracy and discrepancy. 
	From left to right, the number of classes of training data per client increase, which means the degree of data heterogeneity decreases. The discrepancy is defined in Equation~\eqref{for:discrepancy}.}
	\label{fig:motivation}
\end{figure*}

We build three non-IID and one IID MNIST datasets with different class limitations.
The total client size $N=1000$, the number of clients selected in each round $K=20$, the local mini-batch size is $10$, the local epoch $E=20$, the learning rate $\eta=0.03$.
We run $T=300$ rounds of training and evaluated the global model on the local test set.
Note that although we use the same test data in each round, we did not fine-tune the model structure based on the testing results, so we ignore the possibility of test data leakage~\cite{dwork2015reusable}.

The experimental results are shown in Fig.~\ref{fig:motivation}. 
Where the left y-axis is the testing top-1 classification accuracy, and the right y-axis is the arithmetic mean of norm difference between the clients' model weights and the latest global model weights. Specifically, the discrepancy in communication round $t$ is defined as:
\begin{equation}
	\label{for:discrepancy}
	Discrepancy(t) \triangleq \frac{1}{|S_t|} \sum_{i \in S_t}\|\bm{w}_i-\bm{w}_t\|.
\end{equation}

As shown by the blue lines in Fig.~\ref{fig:motivation}, the discrepancy is relaxed as the class number limitation of each client's data increases, which means the high heterogeneity will make the trained models prone to diverge. 
Moreover, high data heterogeneity also hurt the convergence rate of training and the model accuracy, which are shown by the black lines in Fig~\ref{fig:motivation}. The accuracy curves become more fluctuant with the increase of data heterogeneity. 

The quantitative results which are shown in TABLE~\ref{tab:motivation} support our observations. 
As the heterogeneity increases, the variance of the discrepancy significantly decreases by 92.5\% (from 0.11 to 0.0082), and the max accuracy is increased +4.3\% (from 87.9\% to 92.2\%). 
The number of the required communication rounds to reach 85\% accuracy is also significantly reduced in the IID setting, which means faster convergence and less communication consumption.

\begin{table}[]
	\caption{Quantitative results of FedAvg training based on non-IID and IID MNIST with different \#classes/client.}
	\centering
	\label{tab:motivation}
	\footnotesize
	\begin{tabular}{cccccc}
		\hline
		\multicolumn{1}{c|}{\multirow{2}{*}{\# Classes}} & \multicolumn{2}{c|}{Discrepancy} & \multicolumn{2}{c|}{Accuracy} & \# Round to Reach \\ \cline{2-5}
		\multicolumn{1}{c|}{} & \multicolumn{1}{c|}{Mean} & \multicolumn{1}{c|}{Variance} & \multicolumn{1}{c|}{Max} & \multicolumn{1}{c|}{Median} & Target Acc-85\% \\ \hline
		1			&	0.767	&	0.11	& 87.9\% & 79.3\% & 39 \\
		3			&	0.767	&	0.073	& 90.9\% & 87.0\% & 14 \\
		5			&	0.685	&	0.021	& 91.3\% & 90.0\% & 10 \\
		10 (IID)	&	0.627	&	0.0082	& 92.2\% & 91.5\% & 4 \\ \hline
	\end{tabular}
\end{table}

\subsection{Framework Overview}
To tackle the statistical heterogeneity challenge in FL, we propose FedGroup, which groups the training of clients based on the similarities between the client's local optimization solutions.
Our proposed framework is inspired by CFL~\cite{sattler2020clustered}, which clusters clients by a recursive bi-partitioning algorithm. 
CFL assumes the clients have incongruent risk functions (e.g. randomly swapping out the labels of training data), and the goal of clustering is to separate clients with incongruent risk functions.
FedGroup borrows the client clustering idea of CFL.
But unlike previous CFL-based frameworks~\cite{sattler2020clustered, sattler2020byzantine, ghosh2020efficient, xie2020multi, liu2020client}, our grouping strategy is static, which avoids rescheduling clients for each round.
More significantly, we propose the Euclidean distance of Decomposed Cosine similarity (EDC) for efficient clustering of high dimensional direction vectors, which can be regarded as a decomposed variant of MADD~\cite{sarkar2019perfect}.

\begin{figure}[t]
	\centering
	\includegraphics[scale=1.0]{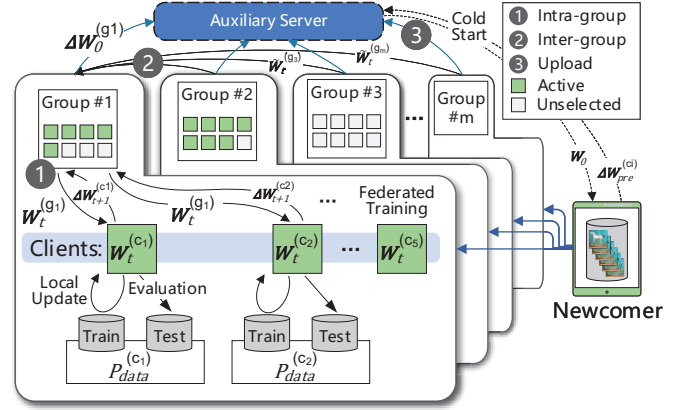}
	\caption{An overview of FedGroup.}
	\label{fig:overview}
\end{figure}

Before we go into more detail, we first show the general overview of the FedGroup. 
The federated training procedure of FedGroup is shown in Fig.~\ref{fig:overview}.
Note that Fig.~\ref{fig:overview} is a schematic diagram, it does not represent the physical deployment of connections or devices.
In practice, the clients can be connected mobiles or IoT devices, the auxiliary server can be deployed in the cloud (FL server), and the group can be deployed on the FL server or the mobile edge computing (MEC) server.
To ease the discussion, we assume that all groups are deployed on the same cloud server, and all clients are connected mobile devices.

FedGroup contains one auxiliary server, a certain amount of groups, and multiple clients, each of them maintains the latest model and the latest update for this model.
Each client and group have a one-to-one correspondence, but there are also possible to have a group without clients in a communication round (empty group) or a client is not in any groups (e.g. a newcomer joins the training later).
The newcomer device uses a cold start algorithm to determine the assigned group (we call the unassigned client is cold), we will explain the details of the cold start algorithm later.
For the dataset, each client $c_i$ has a training set and a test set according to its local data distribution $p_{data}^{(c_i)}$. 

As shown in the Fig.~\ref{fig:overview}, FedGroup has three model transmission processes, including intra-group aggregation (\ding{192}), inter-group aggregation (\ding{193}), and optimization gradient upload (\ding{194}). 
First, the auxiliary server determines the initial model and optimization direction of each group through clustering. 
Then, each group federally trains on a certain set of clients based on their training data to optimize its group model, and evaluates the group model on the same set of clients based on its test data.
Specifically, each group broadcasts its model parameters to their clients and then aggregates the updates from these clients using \textit{FedAvg} in parallel, we call this aggregation procedure the intra-group aggregation.
After all federated trainings of groups complete, the models are aggregated using a certain weight, which we call inter-group aggregation.
Note that, unlike the previous FedAvg-based frameworks~\cite{mcmahan2017communication, smith2017federated, bonawitz2017practical, li2018federated, duan2019astraea}, the optimization gradients in the server is not broadcast to all clients or groups, we only maintain this gradients for the cold start of newcomers.

\textbf{Training:} The details of the training procedure of FedGroup and FedGrouProx (revised by FedProx) are shown in Algorithm~\ref{alg:fedgrouprox}, where $m$ controls the number of groups.
Note that, FedGroup can be regarded as a special case of FedGrouProx with $\mu=0$.
Before training, the server first initializes the global model and group models to the same initial weights $\bm{w}_0$. 
At the beginning of each round, a random subset of clients is selected to participate in this round of training (line 4).
Each group trains the model using FedAvg and get a temporary group model $\tilde{\bm{w}}_{t+1}^{(g_j)}$ (line 7), the group parameters are refreshed after the inter-group aggregation (line 10). 
At the end of each round, the server refreshes the global model by averaging all group models (line 11).
Some details not shown in Algorithm~\ref{alg:fedgrouprox} are that each agent calculates and stores its latest update after refreshing the local model, and the groups and clients need to tackle cold start issues before training. We will discuss our strategies for cold start in the next two sections. 
When $\eta_g = 0$, our framework is pluralistic~\cite{lee2020tornadoaggregate} like the previous CFL works (e.g.~\cite{ghosh2020efficient, sattler2020clustered}).
The inter-group aggregation ($\eta_g > 0$) constructs a novel \textit{semi-pluralistic} architecture for CFL-based frameworks, and we will explore it experimentally in Section~\ref{sec:effects}.

\begin{algorithm}[t]
	\caption{FedGroup and FedGrouProx}
	\label{alg:fedgrouprox}
	\footnotesize
	\begin{flushleft}
	\textbf{Input:} Clients set $\mathcal{C} \leftarrow \{c_1, c_2, ..., c_n\}$, groups set $\mathcal{G} \leftarrow \{g_1, g_2, ..., g_m\}$, \\
	initial group parameters set $\mathcal{W}^{(G)}_0 \leftarrow \{\bm{w}^{(g_1)}_0, \bm{w}^{(g_2)}_0, ..., \bm{w}^{(g_m)}_0\}$, \\
	$g_j.clients \leftarrow \{c_i | c_i \text{~is in group~}g_j, \forall c_i \in \mathcal{C} \}$,
	initial model parameters $\bm{w}_0$, number of communication rounds $T$, number of selected clients per round $K$, inter-group learning rate $\eta_g$, proximal hyperparmeter $\mu$.\\
	\textbf{Output:} Updated group model parameters $\mathcal{W}^{(G)}_T$. 
	\end{flushleft}
	\begin{algorithmic}[1]
		\Procedure {FedGrouProx Training}{}
		\State $\mathcal{W}^{(G)}_1 \leftarrow \{\bm{w}^{(g_1)}_1, \bm{w}^{(g_2)}_1, ..., \bm{w}^{(g_m)}_1\} \leftarrow$ initialized $\mathcal{W}^{(G)}_0$ by $\bm{w}_0$.
		\For{each communication round $t=1,2,...,T$}
		\State $S_t \leftarrow$ Server selects a random subset of $K$ clients.
		\For{each group $g_j$ in $\mathcal{G}$ parallelly}
		\State $S_t^{(g_j)} \leftarrow \{c_i|c_i \in g_j.clients, \forall c_i \in S_t\}$.
		\State $\tilde{\bm{w}}_{t+1}^{(g_j)} \leftarrow$~\textbf{IntraGroupUpdate($S_t^{(g_j)}$, $\bm{w}_{t}^{(g_j)}$)}.
		\State $\tilde{\mathcal{W}}_{t+1}^{(G)} \leftarrow \{\tilde{\bm{w}}_{t+1}^{(g_1)}, \tilde{\bm{w}}_{t+1}^{(g_2)},...,\tilde{\bm{w}}_{t+1}^{(g_m)}\}$.
		\EndFor
		\For{each group $g_j$ in $\mathcal{G}$ parallelly}
		\State $\mathcal{W}_{t+1}^{(G)} \leftarrow$~\textbf{InterGroupAggregation($\tilde{\mathcal{W}}_{t+1}^{(G)}$, $\eta_g$)}.
		\EndFor
		\EndFor
		\EndProcedure

		\Function{IntraGroupUpdate}{$S_t, \bm{w}_t$}
		\If{$S_t$ is $\emptyset$} \Return $\bm{w}_t$.~//Empty group.
		\EndIf
		\State IF \textbf{FedGroup}: $\bm{w}_{t+1} \leftarrow FedAvg(S_t, \bm{w}_t)$.~//Ref.~Algorithm~\ref{alg:fedavg}.
		\State IF \textbf{FedGrouProx}: $\bm{w}_{t+1} \leftarrow FedProx(S_t, \bm{w}_t,\mu)$.~//Ref.~\cite{li2018federated}.
		\State \Return $\bm{w}_{t+1}$
		\EndFunction

		\Function{InterGroupAggregation}{$\tilde{\mathcal{W}}_{t+1}, \eta_g$}
		\For{each group parameter $\tilde{\bm{w}}_{t+1}^{(g_j)}$ in $\tilde{\mathcal{W}}_{t+1}$ parallelly}
		\State  $\Delta\tilde{\bm{w}}_{t+1}^{(g_j)} \leftarrow \eta_g 
		\sum_{l\neq j} \frac{\tilde{\bm{w}}_{t+1}^{(g_l)}}{\|\tilde{\bm{w}}_{t+1}^{(g_l)}\|}$, then $\bm{w}_{t+1}^{(g_j)} \leftarrow \tilde{\bm{w}}_{t+1}^{(g_j)} + \Delta\tilde{\bm{w}}_{t+1}^{(g_j)}$.
		\EndFor
		\State \Return $\mathcal{W}_{t+1} \leftarrow \{\bm{w}^{(g_1)}_{t+1}, \bm{w}^{(g_2)}_{t+1}, ..., \bm{w}^{(g_m)}_{t+1}\}$
		\EndFunction
	\end{algorithmic}
\end{algorithm}
 
\subsection{Group Cold Start}
In the general setting of federated learning, there is only one global optimization goal, which is to minimize the joint loss function of all learning tasks of all clients.
However, optimizing a global objective becomes difficult when the local data distributions are non-IID or called statistical heterogeneity.
Li \textit{et al.}~\cite{Li2020On} theoretically prove that the heterogeneity will also slow down the convergence rate. 
Instead of optimizing a complex global goal, why not divide it into several sub-optimization goals? 
Based on this idea, we manage to group the trainings of clients based on the proximities between their local optimizations.
But there are two following questions: (1) how to measure the proximity or distance, and (2) how to determine the optimization goals of each group before training.

For the choice of measure, a heuristic way is to use the $\ell_2$ distance between models, which has miserable performance as we mentioned in \ref{sec:background-CLF}.
The loss value can be used as a surrogate for the proximity of difference domains~\cite{mohri2019agnostic}, but the huge computational overhead involved in model inferencing.
The cosines similarity between the gradients calculated by backpropagation or the updates of model parameters is an alternative measure.
The cosine similarity between the updates of any two clients $c_i$ and $c_j$ is defined by:
\begin{equation}
	\mathcal{S}(i,j) \triangleq \frac{<\Delta\bm{w}_t^{(c_i)}, \Delta\bm{w}_t^{(c_j)}>}{\|\Delta\bm{w}_t^{(c_i)}\|~\|\Delta\bm{w}_t^{(c_j)}\|},
\end{equation}
and the pairwise cosine similarity matrix $\mathcal{M} \subset \mathbb{R}^{n \times n}$ can be written with cosine similarity kernel $K$ as follows:

\begin{equation}\label{eq:cossim}
	\mathcal{M} = K(\Delta \bm{W},\Delta \bm{W}),~\mathcal{M}_{ij} = \mathcal{S}(i,j).
\end{equation}

The computational complexity of calculating $\mathcal{M}$ is $O\big(n^2d^2_{\bm{w}}\big)$, $n$ and $d_{\bm{w}}$ are the number of clients and number of parameters, respectively.
We assume the all parameters updates $\Delta\bm{w}_t$ are flattened row vectors, so $\Delta\bm{w}_t \subset \mathbb{R}^{1 \times d_{\bm{w}}}$.
In general, the number of parameters $d_{\bm{w}}$ is huge and $d_{\bm{w}} \gg n$ (HDLSS), which make the pairwise cosine similarity-based clustering methods computationally inefficient.
Unlike $\ell_p$ distance, the expectation of $\mathcal{S}$ asymptotically remains constant as dimensionality increases~\cite{radovanovic2010existence}, which is friendly to the clustering in high dimensional data.
Unfortunately, $\mathcal{S}$ is not suitable for low-dimensional situations because the variance of it is $O(1/d_{\bm{w}})$.
Therefore, we extend the data-driven method MADD~\cite{sarkar2019perfect} to our cosine similarity-based case, so we can reduce the observation bias by using the mean of residuals of $\mathcal{S}$, fox example:

\begin{equation}\label{eq:MADC}
	MADC(i,j)=\frac{1}{n-2} \sum_{z \neq i,j} |\mathcal{S}(i, z) - \mathcal{S}(j, z)|.
\end{equation}

The above dissimilarity measure is based on the Mean of Absolute Differences of pairwise Cosine similarity, so we call it MADC.
However, MADC and $\mathcal{M}$ have the same computation complexity and both are proximity measures, which means they cannot be applied for the efficient Euclidean distance-based clustering algorithms. These motivate us to develop a variant of MADC called Euclidean distance of Decomposed Cosine similarity (EDC), which is defined by:

\begin{equation}\label{eq:EDC}
	\begin{aligned}
	EDC(i,j)=&~\frac{1}{m} \sqrt{\sum_{v \in \bm{V}} {(\mathcal{S}(i, v) - \mathcal{S}(j, v))}^2}, ~or\\
	EDC(i,j)=&~\frac{1}{m} \| K(\Delta\bm{w}_t^{(c_i)}, \bm{V}^T) - K(\Delta\bm{w}_t^{(c_j)}, \bm{V}^T)\|, \\
	\bm{V}=&~SVD(\Delta \bm{W}^T, m), \bm{V} \subset \mathbb{R}^{d_{\bm{w}} \times m}.
	\end{aligned}
\end{equation}

Instead of calculating the pairwise similarity, EDC first decomposes the updates of models into $m$ directions by using truncated Singular Value Decomposition (SVD) algorithm~\cite{golub1971singular}, so then only the similarities between the updates and these directions will be calculated. It is worth noting that the complexity of truncated SVD is only $O(2m^2d_{\bm{w}})$ for $d_{\bm{w}} \gg m$ and hence the computational complexity of EDC is $O(m^2d_{\bm{w}}^2)$. Some previous works (i.e.~\cite{sattler2020clustered, briggs2020federated}) calculate the pairwise cosine similarity based on all participants, so $n$ is usually hundreds or thousands. Furthermore, \cite{sattler2020clustered, briggs2020federated} leverage the hierarchical clustering strategies, which are recursively and computationally expensive.

To determine the optimization goals of each group, FedGroup clusters the parameter updates directions of clients into $m$ groups using K-Means++~\cite{arthur2006k} algorithm based on EDC.
The main advantage of the clustering approach is that it is unsupervised, and we can divide the global optimization function into $m$ sub-optimization functions regardless of whether there have incongruent optimization goals.
Moreover, training all clients to cluster their updated directions is not communication-friendly and practically achievable.
In other words, FedGroup performs a low-dimensional embedding of the local updates matrix $\Delta \bm{W}$, following by K-Means++ clustering.
Our calculation of client clustering is following the calculation of the similarity matrix, which means the above calculations only require one round of communication.
We call the combination of the above two processes as group cold start, and the details are shown in Algorithm~\ref{alg:groupcoldstart}. For comparison, we also provide the MADC version of FedGroup, which is clustered using the hierarchical strategy with the complete linkage.

After the server performs the group cold start, the optimization direction of group $j$, which is measured by $\Delta \bm{w}_0^{(g_j)}$ is determined, and the $\alpha m$ clients participating in this process are assigned. 
We leverage the centric means of groups to measure the clustering validity index like within-cluster sum-of-squares criterion.
The optimization gradients will be uploaded to the auxiliary server for cold start of the newcomers.

\begin{algorithm}[t]
	\caption{Group Cold Start}\label{alg:groupcoldstart}
	\footnotesize
	\begin{flushleft}
		\textbf{Input:} Clients set $\mathcal{C}$, number of group $m$, global initial model $\bm{w}_0$, pre-training scale hyperparameter $\alpha$. \\
		\textbf{Output:} Groups set $\mathcal{G}$, set of group parameters $\mathcal{W}_0^{(G)}$, set of group updates $\Delta \mathcal{W}_0^{(G)}$.
	\end{flushleft}
	\begin{algorithmic}[1]
		\Procedure {Group Cold Start}{}
		\State $S \leftarrow$ Server selects a random subset of $\alpha*m$ clients.
		\State Server broadcasts $\bm{w}_0$ to all selected clients.
		\For{each client $c_i$ in $S$ parallelly}
		\State $\Delta \bm{w}_0^{(c_i)} \leftarrow$~$flatten($ \textbf{ClientUpdate($i, \bm{w}$)} $)$.~//Ref.~Algorithm~\ref{alg:fedavg}.
		\EndFor
		\State $\Delta \bm{W} \leftarrow [\Delta \bm{w}_0^{(c_1)}, \Delta \bm{w}_0^{(c_2)}, \dots, \Delta \bm{w}_0^{(c_{\alpha*m})}]$.
		
		\If{MADC}:
		\State $\mathcal{M} \leftarrow K(\Delta \bm{W},\Delta \bm{W})$.~//Ref.~Eq.~(\ref{eq:cossim})
		\State Proximity matrix $\mathcal{M}_p \leftarrow$ \textbf{Calculate MADC($\mathcal{M}$)}.~//Ref.~Eq.~(\ref{eq:MADC})
		\State $[g_1.clients, \dots, g_m.clients] \leftarrow$ Hierarchical Clustering($\mathcal{M}_p, m$).
		\EndIf
		\If{EMD}:
		\State $V \leftarrow truncated~SVD(\Delta \bm{W}^T,m)$.
		\State Distance matrix $\mathcal{M}_d \leftarrow K(\Delta \bm{W}, V^T)$.~//Ref. ~Eq.~(\ref{eq:EDC})
		\State $[g_1.clients, \dots, g_m.clients] \leftarrow$ K-Means++($\mathcal{M}_d,m$).
		\EndIf
		
		\For{$g_j$ in $\mathcal{G} \leftarrow [g_1, \dots, g_m]$}
		\State $\bm{w}_0^{(g_j)} \leftarrow Average([\Delta \bm{w}_0^{(c_i)},\forall c_i \in g_j.clients])$.
		\State $\Delta \bm{w}_0^{(g_j)} \leftarrow \bm{w}_0^{(g_j)} - \bm{w}_0$.
		\EndFor
		\State $\mathcal{W}_0^{(G)} \leftarrow [\bm{w}_0^{(g_1)}, \dots, \bm{w}_0^{(g_m)}]$, and $\Delta \mathcal{W}_0^{(G)} \leftarrow [\Delta \bm{w}_0^{(g_1)}, \dots, \Delta \bm{w}_0^{(g_m)}]$.
		\State \Return $\mathcal{G}, \mathcal{W}_0^{(G)}, \Delta \mathcal{W}_0^{(G)}$
		\EndProcedure
	\end{algorithmic}
\end{algorithm}

Another improvement of our algorithm is that we only select a subset of clients to participate in the pre-training and decomposition process. 
We control the scale of pre-training by hyperparameter $\alpha$ and the number of pre-training clients is set to $\alpha m$.
Our implementation is more practical because it is difficult to satisfy that all clients are active until they complete the training in the large-scale FL systems. For example, the drop-out may occur due to the network jitter.
We will show the efficiency of EDC in Section~\ref{sec:evaluation}.

\subsection{Client Cold Start}
As described before, the group cold start algorithm selects a random subset ($\alpha m $) of the clients for pre-training, so the remaining clients ($n-\alpha m$) are cold clients and are not in any groups.
Since the federated training network is dynamic, the new devices can join the training at any time, so we need to classify newcomers according to the similarity between their optimization goals and groups'.
Our client cold start strategy is to assign clients to the groups that are most closely related to their optimization direction, as shown below:
\begin{equation}
	\begin{aligned}
	\label{for:clientcoldstart}
	&g^* = \operatorname*{argmin}_j \frac{-\cos(\sphericalangle(\Delta \bm{w}_0^{(g_j)}, \Delta \bm{w}_{pre}^{(i)}))+1}{2}. \\
	\end{aligned}
\end{equation}
Suppose the newcomer $i$ joins the training network in round t, then the $\Delta \bm{w}_{pre}^{(i)}$ is the pre-training gradient of newcomer base on the global initial model $\bm{w}_0$.
We schedule the newcomer $i$ to group $g^*$ to minimize the normalized cosine dissimilarity.
With this mechanism, FedGroup does not need to broadcast all groups' models for clustering.

In summary, the key features of our FedGroup framework are as follows:
\begin{itemize}
	\item FedGroup reduces the discrepancy between the joint optimization objective and sub-optimization objectives, which will help improve the convergence speed and performance of federated learning systems.
	\item The proposed framework determines the optimization objectives of groups by a clustering approach, which is unsupervised and can disengage from the incongruent risk functions assumption.
	\item FedGroup adopts a novel decomposed data-driven measure named EDC, a low-dimension embedding of clients' local updates, for efficient HDLSS direction vectors clustering.
	\item The clustering mechanism of FedGroup considers the joining of newcomer devices.
\end{itemize}

Compared with vanilla FL, FedGroup requires additional communication to transmit pre-training parameters. 
However, we emphasize that the pre-training procedure does not occupy a whole round, the client can continue to train $E-1$ epochs and upload the parameters updates for the intra-group aggregation.

\section{Evaluation}\label{sec:evaluation}
In this section, we present the experimental results for FedGroup and FedGrouProx frameworks.
We show the performance improvements of our frameworks on four open datasets.
Then we demonstrate the effectiveness of our grouping strategy, which includes the clustering algorithm (group cold start) and the newcomer assignment algorithm (client cold start).
We further study the relationship between the EDC and MADC.

Our implementation is based on Tensorflow~\cite{abadi2016tensorflow}, and all code and data are publicly available at \textit{github.com/morningD/GrouProx}.
To ensure reproducibility, we fix the random seeds of the clients' selection and initialization.

\subsection{Experimental Setup}
We evaluate FedGroup and FedGrouProx on four federated datasets, which including two image classification tasks, a synthetic dataset, a sentiment analysis task. 
In this section, we adopt the same notation for federated learning settings as Section~\ref{sec:design} and as~\cite{li2018federated}: the local epoch $E=20$, the number of selected clients per round $K=20$, the pre-training scale $\alpha=20$, the inter-group aggregation is disable. The local solver is mini-batch SGD with $B=10$. Besides, the learning rate $\eta$ and FedProx hyperparameter $\mu$ in our experiments are consistent with the recommended settings of~\cite{li2018federated}. 

\noindent\textbf{Datasets and Models}
\begin{itemize}
\item MNIST~\cite{lecun1998gradient}. A 10-class handwritten digits image classification task, which is divided into 1,000 clients, each with only two classes of digits. We train a convex multinomial logistic regression (MCLR) model and a non-convex multilayer perceptron (MLP) model based on it. The MLP has one hidden layer with 128 hidden units.
\item Federated Extended MNIST (FEMNIST)~\cite{cohen2017emnist}. A 62-class handwritten digits and characters image classification task, which is built by resampling the EMNIST~\cite{cohen2017emnist} according to the writer. Similar to the experiment of EMNIST, we train a MCLR model and a MLP model (one hidden layer with 512 hidden units) based on it.
\item Synthetic. It's a synthetic federated dataset proposed by \textit{Shamir et. al}~\cite{shamir2014communication}. Our hyperparameter settings of this data-generated algorithm are $\alpha=1, \beta=1$, which control the statistical heterogeneity among clients. We study a MCLR model based on it.
\item Sentiment140 (Sent140)~\cite{go2009twitter}. A tweets sentiment analysis task, which contains 772 clients, each client is a different twitter account. We explore a LSTM classifier based on it.
\end{itemize} 
The statistics of our experimental datasets and models are summarized in TABLE~\ref{tab:dataset}.

\begin{table}[h]                                                         
	\caption{Statistics of Federated Datasets and Models.}                 
	\centering
	\small
	\label{tab:dataset}
	\begin{tabular}{lllll}
		\hline
		Dataset & Devices & Samples & Model & \# of Parameters \\ \hline
		\multirow{2}{*}{MNIST} & \multirow{2}{*}{1,000} & \multirow{2}{*}{69,035} & MCLR & 7,850 \\ \cline{4-5} 
		&                   &                   & MLP & 101,770 \\ \hline 
		\multirow{2}{*}{FEMNIST} & \multirow{2}{*}{200} & \multirow{2}{*}{18,345} & MCLR & 20,410 \\ \cline{4-5} 
		&                   &                   & MLP & 415,258 \\ \hline 
		Synthetic(1,1)& 100 & 75,349 & MCLR & 610 \\ \hline
		Sent140 & 772 & 40,783 & LSTM & 243,861 \\ \hline
	\end{tabular}
\end{table}

\noindent\textbf{Baselines}
\begin{itemize}
\item FedAvg~\cite{mcmahan2017communication}: the vanilla federated learning framework.

\item FedProx~\cite{li2018federated}: the popular federated learning optimizer toward heterogeneous networks.

\item IFCA~\cite{ghosh2020efficient}: An CFL framework that minimizes the loss functions while estimating the cluster identities.

\item FeSEM~\cite{xie2020multi}: An $\ell_2$ distance-based CFL framework that minimizes the expectation of discrepancies between clients and groups stochastically.
\end{itemize} 

\begin{table*}[t]
	\centering
	\scriptsize
	\caption{Comparisons with FedAvg, FedProx, IFCA, FeSEM, FedGroup (FG), FedGrouProx (FGP) on MNIST, FEMNIST, Synthetic, Sent140. Ablation studies of FedGroup: Random Cluster Centers (RCC), Randomly Assign Cold (i.e. newcomers) Clients (RAC). The accuracy improvements $\uparrow$ are calculated relative to the $\ell_2$ distance-based CFL framework FeSEM.}
	\begin{tabular}{|l|cc|cc|cc|cccc|}
		\hline
		\multirow{2}{*}{Dataset-Model} & FedAvg & FedProx & IFCA & FeSEM & FG-RCC- & FG-RAC- & FG- & FGP- & FG- & FGP- \\
		& \cite{mcmahan2017communication} & \cite{li2018federated} & \cite{ghosh2020efficient} & \cite{xie2020multi} & EDC & EDC & EDC & EDC & MADC & MADC \\ \hline
		MNIST-MCLR & $89.8$ & $90.7$ & $95.3$ & $89.1$ & $92.6$ & $88.8$ & $96.0(\uparrow6.9)$ & $\bm{96.1}(\uparrow7.0)$ & $95.6(\uparrow6.5)$ & $95.3(\uparrow6.2)$ \\
		MNIST-MLP & $95.3$ & $94.6$ & $97.8$ & $93.9$ & $96.5$ & $94.3$ & $\bm{97.9}(\uparrow4.0)$ & $97.7(\uparrow3.8)$ & $\bm{97.9}(\uparrow4.0)$ & $97.7(\uparrow3.8)$ \\
		FEMNIST-MCLR & $78.8$ & $77.9$ & $87.9$ & $65.6$ & $88.8$ & $74.7$ & $\bm{91.7}(\uparrow26.1)$ & $\bm{91.7}(\uparrow26.1)$ & $91.2(\uparrow25.6)$ & $90.7(\uparrow25.1)$ \\
		FEMNIST-MLP & $82.7$ & $81.0$ & $95.5$ & $69.9$ & $91.5$ & $78.4$ & $\bm{96.8}(\uparrow26.9)$ & $95.4(\uparrow25.5)$ & $94.6(\uparrow24.7)$ & $93.6(\uparrow23.7)$ \\
		Synthetic(1,1)-MCLR & $74.7$ & $81.5$ & $\bm{95.5}$ & $92.1$ & $92.9$ & --- & $88.4(\downarrow3.7)$ & $87.4(\downarrow4.7)$ & $92.1(=)$ & $88.5(\downarrow3.6)$ \\
		Sent140-LSTM & $72.2$ & $71.9$ & $\bm{76.8}$ & $70.0$ & $76.1$ & $72.2$ & $75.3(\uparrow5.3)$ & $75.4(\uparrow5.4)$ & $75.8(\uparrow5.8)$ & $75.4(\uparrow5.4)$ \\ \hline
	\end{tabular}
	\label{tab:result}
\end{table*}

\noindent\textbf{Evaluation Metrics} 

Since each client has a local test set in our experimental setting, we evaluate its corresponding group model based on these data. 
For example, in FedAvg and FedProx we evaluate the global model based on the test set for all clients. 
And in FedGroup and FedGrouProx we evaluate the group model based on the test set for the clients in this group.
We use top-1 classification accuracy to measure the performance of the classifiers.
Given that FedGroup has multiple accuracies of groups with different sizes, we use a "weighted" accuracy to measure the overall performance and the weight is proportional to the test data size of each group.
In fact, the "weighted" accuracy is equivalent to the sum of the misclassified sample count in all groups divided by the total test size.

To make our results more comparable, the test clients of the group model is all the clients historically assigned to this group. 
As the number of training rounds increases, all clients would be included in the test.
Note that the heterogeneity will affect the convergence, resulting in greater fluctuations in model accuracy during the training process.
Therefore, we assume the early stopping~\cite{yao2007early} strategy is applied and we regard the maximum test accuracy during training as the final score. The number of groups of all CFL-based frameworks remains the same for each dataset. 

\begin{figure*}[t]
	\centering
	\subfloat[MNIST-MCLR]{
		\hspace{-8mm}
		\includegraphics{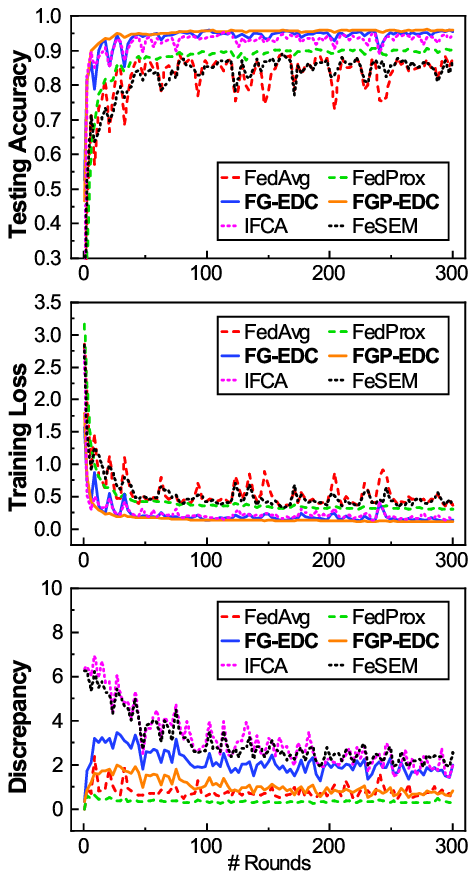}
		\label{fig:mnist1}
		\hspace{-8mm}
	}
	\subfloat[MNIST-MLP]{
		\includegraphics{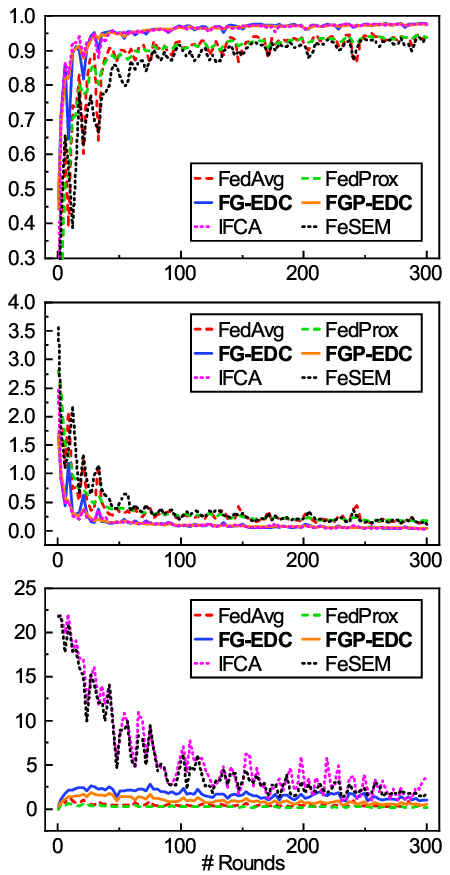}
		\label{fig:mnist2}
		\hspace{-8mm}
	}
	\subfloat[FEMNIST-MCLR]{
		\includegraphics{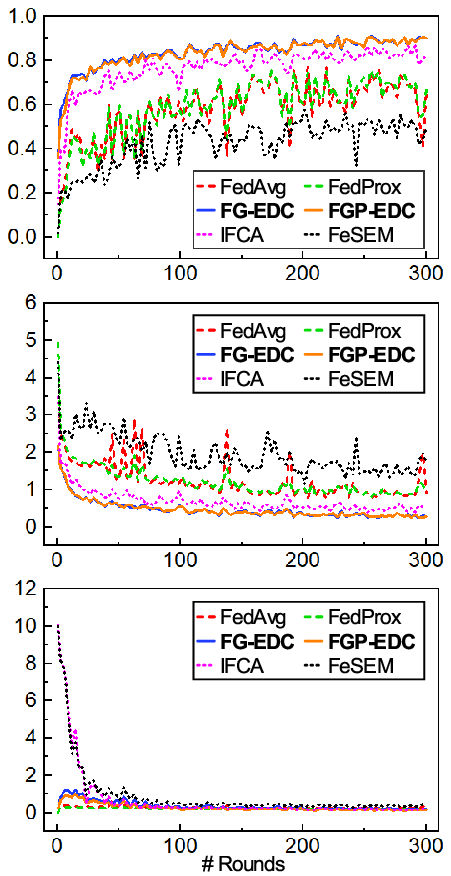}
		\label{fig:nist1}
		\hspace{-8mm}
	}
	\subfloat[FEMNIST-MLP]{
		\includegraphics{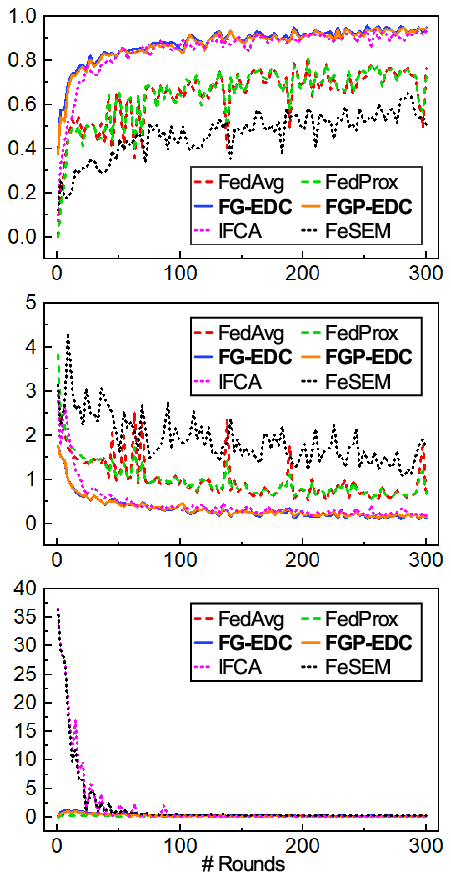}
		\label{fig:nist2}
		\hspace{-8mm}
	}
	\caption{Evaluation results on MNIST ($m=3$) and FEMNIST($m=5$). Top: test accuracy; Middle: mean training loss; Bottom: discrepancy between clients and server (FedAvg and FedProx) or weighted discrepancy between clients and groups (FedGroup, FedGrouProx, IFCA, FeSEM).}
	\label{fig:acc1}
	\vspace{-4mm}
\end{figure*}

\begin{figure}[t]
	\centering
	\subfloat[Synthetic(1,1)-MCLR]{
		\hspace{-3mm}
		\includegraphics{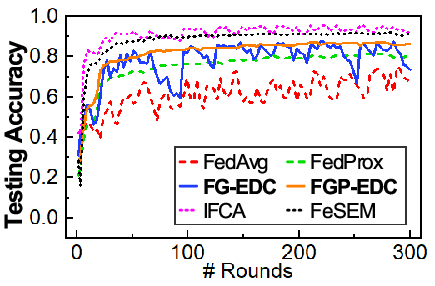}
		\label{fig:syn}
	}
	\subfloat[Sent140-LSTM]{
		\hspace{-3mm}
		\includegraphics{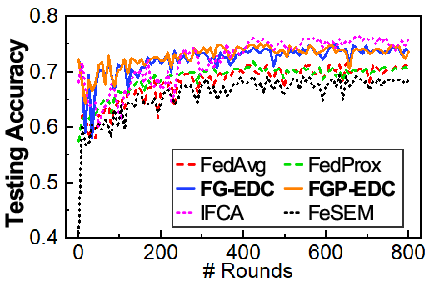}
		\label{fig:lstm}
	}
	\caption{Classification accuracy evaluated on Synthetic(1,1) ($m=5$) and Sent140 ($m=5$).}
	\label{fig:acc2}
\end{figure}

\begin{figure}[]
	\centering
	\subfloat[FEMNIST-MLP]{
		\hspace{-3mm}
		\includegraphics{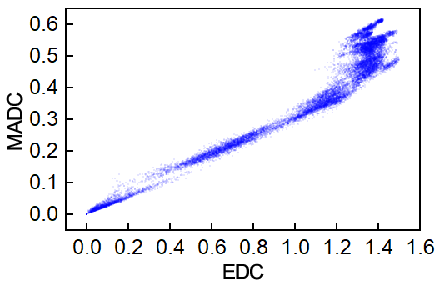}
	}
	\centering
	\subfloat[Sent140-LSTM]{
		\hspace{-3mm}
		\includegraphics{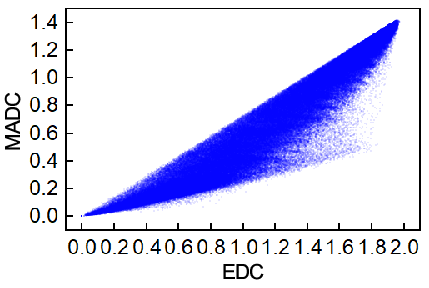}
	}
	\caption{The mapping from the MADC to the EDC, the pre-training scale $\alpha=20$.}
	\label{fig:sim}
\end{figure}

\subsection{Effects of Proposed Framework}
\label{sec:effects}

In TABLE~\ref{tab:result}, we compare the evaluated results of FedGroup (FG) and FedGrouProx (FGP) with baselines.
We calculate the accuracy improvements related to the FeSEM.
The results show that FG, FGP, IFCA are significantly superior to other frameworks.
In particular, FG-EDC improves absolute test accuracy by $+26.9\%$ on FEMNIST, which is highly statistical heterogeneity.
The FeSEM performs worst in all datasets expect Synthetic(1,1), which can be interpreted as the lower dimensionality ($d_{\bm{w}}=610$) of model.
The EDC is mostly slightly better than MADC, but this gap is shortened in the low dimensions cases, especially Synthetic(1,1).
To investigate the effects of our proposed strategies, we further perform two ablation studies: RCC (random cluster centers) and RAC (randomly assign cold clients, but the pre-training clients are retained in their groups).
In the RCC setting, the accuracy is moderately degraded (expect Synthetic) but still surpasses the FeSEM.
The RAC strategy leads to a significant decrease in accuracy ($-7.86\%$ on average) and the final scores are even worse than the FedAvg.
Therefore, the combination of our clustering algorithm and client cold start strategy is efficient and can achieve more performance improvements.

The experimental results in specific rounds are shown in Fig.~\ref{fig:acc1} and Fig.~\ref{fig:acc2}.
The accuracy and discrepancy curves illustrate that adding the proximal term can reduce the divergence caused by the heterogeneous data and make the training more stable.
Although the training procedure of FedGrouProx is more stable than that of FedGroup, there is not enough evidence to suggest that adding the proximal term is significantly helpful in improving accuracy.
Compare to IFCA and FeSEM, the discrepancy of FedGroup remains at a low level in the initial round, which is attributed to the effective group cold start strategy. 
The FedAvg and FedProx achieve low discrepancy by maintaining the consensus global model, but this will degrade the accuracy, which is supported by the obtained results.
FedGroup also shows a significant improvement in convergence speed compared to FedAvg, which is helpful to reduce the communication consumption of the FL systems.
By the way, the rescheduling methods of IFCA and FeSEM involve additional overheads in each round, such as broadcasting group models and calculating loss.


To explore the potential of semi-pluralistic architecture, we evaluate FedGroup under difference $\eta_g$. 
The details are not presented due to the pages-limit.
Our experiments in MNIST-MCLR and Sent140-LSTM show that the inter-group aggregation mechanism can slightly improve the convergence rate of model training.
Unfortunately, the convergence rates of other experimental sets do not improve as expected, which can be interpreted as each group in FedGroup is highly specialized and has few common representations. So, we will explore using a gate network to combine group models in the future.


Finally, in Fig.~\ref{fig:sim}, we study the difference between MADC and our decomposed measure EDC as clustering distance.
The experiments show that the mapping from the MADC to EDC is approximately linear, which means that EDC can be regarded as a low-dimensional linear approximation of MADC.
In addition, calculating EDC only needs to pre-train $\alpha m$ clients, which is more computationally efficient and implementable than calculating pairwise proximity.

\section{Conclusion}\label{sec:conclusion}
In this work, we have presented the decomposed data-driven measure-based clustered federated learning frameworks FedGroup, which can improve the model performance of federated training by efficient clustering and cold start strategies.
We evaluated the proposed frameworks on four open datasets and shown the superiority of FedGroup compared to FedAvg, FedProx, FeSEM.
FedGroup significantly improved $+26.9\%$ top-1 test accuracy on FEMNIST compared to FedSEM.
Besides, our experiments on FedGrouProx have found that adding the proximal term can make the federated training more stable, but it does not significantly help to improve classification accuracy. 
The experiments of our novel data-driven measure EDC, shown that it is an appropriate linear approximation of MADC.
Our evaluation results on 6 models shown that FedGroup achieved higher classification accuracy compared to FedAvg, FedProx, and two random strategies RCC (except Sent140) and RAC.

\bibliographystyle{IEEEtran}
\balance
\bibliography{IEEEabrv,references}

\end{document}